\documentclass[runningheads]{llncs}
\usepackage[T1]{fontenc}
\usepackage{graphicx}
\usepackage{adjustbox}
\usepackage{amsmath}
\usepackage{amssymb}
\usepackage{subcaption}
\usepackage{siunitx}

\begin{document}
\title{Latent Space Representation of Electricity Market Curves: Maintaining Structural Integrity}
\titlerunning{Latent Space Repr. of El. Market Curves}
% If the paper title is too long for the running head, you can set
% an abbreviated paper title here
%
\author{
Martin V\'yboh\inst{1}\orcidID{0009-0008-6198-5591} \and
Zuzana Chladn\'a\inst{2}\orcidID{0000-0001-6837-5376} \and
Gabriela Grmanov\'a\inst{1}\orcidID{0000-0001-9024-8584} \and
M\'aria Luck\'a\inst{1}\orcidID{0000-0003-2061-1194}
}
\authorrunning{M. V\'yboh et al.}
% First names are abbreviated in the running head.
% If there are more than two authors, 'et al.' is used.
%
\institute{
Kempelen Institute of Intelligent Technologies, Sky Park Offices Bottova 7939/2A, Bratislava, 811 09, Slovakia \\
\email{\{name.surname\}@kinit.sk}\and
Department of Applied Mathematics and Statistics, Faculty of Mathematics, Physics and Informatics, Comenius University, Mlynska dolina F1, Bratislava, 842 48, Slovakia \\
\email{chladna@fmph.uniba.sk}
}
\maketitle              % typeset the header of the contribution
\begin{abstract}
Efficiently representing supply and demand curves is vital for energy market analysis and downstream modelling; however, dimensionality reduction often produces reconstructions that violate fundamental economic principles such as monotonicity. This paper evaluates the performance of PCA, Kernel PCA, UMAP, and AutoEncoder across 2d and 3d latent spaces. During preprocessing, we transform the original data to achieve a unified structure, mitigate outlier effects, and focus on critical curve segments. To ensure theoretical validity, we integrate Isotonic Regression as an optional post-processing step to enforce monotonic constraints on reconstructed outputs. Results from a three-year hourly MIBEL dataset demonstrate that the non-linear technique UMAP consistently outperforms other methods, securing the top rank across multiple error metrics. Furthermore, Isotonic Regression serves as a crucial corrective layer, significantly reducing error and restoring physical validity for several methods. We argue that UMAP’s local structure preservation, combined with intelligent post-processing, provides a robust foundation for downstream tasks such as forecasting, classification, and clustering.

\keywords{Electricity market  \and Market curves \and Latent space \and Machine learning.}
\end{abstract}

\vspace*{-2em}
\section{Introduction}
\subsection{Related Work}
Despite decades of market liberalisation, predicting electricity prices remains a complex challenge due to high volatility and limited storage options. While traditional approaches have focused on direct price regression \cite{jarka,Ferreira} or standard machine learning algorithms \cite{Lago}, recent research has shifted toward modelling entire supply and demand curves. These curves offer insights exceeding simple market-clearing price analysis, uncovering shifts in production or demand.

Previous studies have primarily concentrated on the supply side, employing techniques such as multivariate linear regression \cite{Ziel}, radial basis function (RBF) \cite{Soloviova}, and piecewise linear functions \cite{yildirim}. Notably, \cite{Guo} and \cite{Tang} addressed the high dimensionality of supply curves by applying dimensionality reduction methods (e.g., PCA) combined with deep learning for forecasting. More recently, \cite{sinha} introduced a modular framework that uses a monotonic AutoEncoder to preserve the monotonicity of reconstructed supply and demand curves.

\vspace*{-0.75em}

\subsection{Our Contribution}
In this work, we propose a framework that builds upon and extends earlier contributions in the field. We simplify and expand the preprocessing approach presented in \cite{Guo}. We introduce a modular component that ensures the monotonicity of curves reconstructed from any latent space. To the best of our knowledge, this is the first study to analyse the ability of various dimensionality reduction methods to capture the intrinsic properties of electricity market curves. It also appears to be the first time that methods such as Kernel PCA and UMAP have been explored for electricity market curve analysis. 

To address the high-dimensional nature of market curves, we compare four distinct reduction techniques: Principal Component Analysis (PCA) for linear transformation \cite{Pearson}, Kernel PCA (kPCA) for non-linear mapping in Hilbert space \cite{Scholkopf}, Uniform Manifold Approximation and Projection (UMAP) for scalable manifold learning \cite{McInnes}, and AutoEncoder (AE) for capturing complex non-linearities through deep learning \cite{wang}. We validate our approach using data from the Iberian Electricity Market.

\vspace*{-1em}
\section{Methodology} \label{sec:Methodology}
\vspace*{-0.5em}
This section describes the framework for transforming supply and demand curves into a latent space. The presented procedure consists of the following phases: data processing and transformation, dimensionality reduction, and post-processing. Each phase is applied independently to both supply and demand curves. 

\vspace*{-0.5em}

\subsection{Data Preprocessing}
Let the dataset consist of $K$ stepwise market curves, where the $i$-th curve is a sequence of price-volume pairs $[P_{i1}^0, Q_{i1}^0], [P_{i2}^0, Q_{i2}^0],\dots,[P_{iN_i}^0, Q_{iN_i}^0],$ where $N_i$ is the number of price steps of curve $i$.

Inspired by the work of \cite{Guo}, we implement a four-step preprocessing pipeline but omit their sigmoid transformation, as it is redundant under standardisation and complicates inverse mapping (produces values outside $[0,1]$ interval):
%\vspace*{-1em}
\begin{enumerate}
    \item \textbf{Price winsorization:} To focus on the price-sensitive region, all prices are winsorized \cite{wilcox} to a 99\% confidence interval $[P_{\min},P_{\max}]$ of historical market-clearing prices. The price-winsorized price-volume pairs are now denoted $[P_{ij}^1, Q_{ij}^1]$. The values of the volume and the number of pairs describing the $i$-th individual curve remain unchanged, i.e. $Q_{ij}^1 = Q_{ij}^0$.
    \item \textbf{Merging in prices:} Here, we solve the problem of non-uniform prices, which represent the individual market curves. We adopt the merging procedure proposed by \cite{Guo}. We map all prices to a uniform grid $R_P$ with a fixed integer step size $\delta_P$. We define $R_P$ as the sequence of integers $\{P_{\text{start}}, P_{\text{start}} + \delta_P, \dots, P_{\text{end}}\}$ that encompasses the winsorized interval $[P_{\min}, P_{\max}]$. Each price from $\{P_{ij}^1\}_{j=1,\dots,N_i}$ is transformed to its nearest neighbour in $R_p$, resulting in discretized series $\{P_{ij}^2\}_{j=1,\dots,N_i}$. This ensures consistent prices across the dataset, while maintaining original volumes, i.e. $Q_{ij}^2 = Q_{ij}^1$.
    \item \textbf{Sampling in volume:}  We decrease the dimension of the market curve representation by uniform sampling of reference bidding prices. This can be done by sampling the reference prices at fixed electricity volume points. Departing from \cite{Guo}, we restrict the volume range to a uniform sequence $R_Q = \{Q_{\text{start}}, Q_{\text{start}} + \delta_Q, \dots, Q_{\text{end}} \}$, that encompasses a 99\% confidence interval (CI) of historical clearing volumes, rather than considering the entire volume range from $Q_{\min}$ to $Q_{\max}$. This shifts the focus to the high-probability region where supply and demand curves intersect. Each curve is resampled onto $R_Q$, resulting in a uniform $K \times N$ price matrix, where $N = |R_Q|$ represents the fixed number of bidding reference prices~$\mathbf{P}^3 =\left\{ P_{ij}^3 \mid i = 1, \dots, K; \, j = 1, \dots, N \right\}.$
    \item \textbf{Standardisation of prices.} Finally, the matrix $\mathbf{P}^3$ is standardised to zero mean and unit variance per dimension (calculated on the training set), yielding the input matrix $\mathbf{P}^4$.
\end{enumerate} 

\vspace*{-1.5em}

\subsection{Dimensionality Reduction}
The high dimensionality of market curves poses a challenge for accurate prediction. However, since these curves often exhibit recurring patterns, dimensionality reduction (DR) techniques can effectively project them into a low-dimensional latent space while preserving their essential characteristics. To achieve this, we employ four methods: PCA, kPCA, UMAP, and AE.

Each method requires hyperparameter tuning, which we perform using two data subsets: a training set for optimisation and a validation set for evaluation. The selected techniques are applied to the $\mathbf{P}^4$ matrix. During evaluation, the validation set is projected into the latent space and then reconstructed back to the original space. The performance of each method is assessed using Root Mean Squared Error (RMSE) and Mean Absolute Error (MAE) of the reconstructed curves in $\mathbf{P}^3$ form.

\vspace*{-1em}

\subsection{Post-Processing}
\vspace*{-0.5em}
Most DR methods do not inherently enforce monotonicity -- a critical property of market curves. To address this, we apply Isotonic Regression (IR) \cite{barlow} to the reconstructed curves, ensuring monotonicity without altering the underlying architecture of the DR methods. The IR model is fitted to all samples from the training set per curve type.

\vspace*{-1em}
\section{Experiments}
\vspace*{-0.5em}
We evaluate the proposed framework using hourly supply and demand curves from the Iberian Electricity Market (MIBEL) \cite{OMIEData} for the period 2018–2020. The dataset ($K = 26\,298$) was divided into training, validation, and test sets using a 7:1:4 ratio (1 year and 9 months : 3 months : 1 year). 

Experiments were implemented in the Python programming language. For DR methods, we utilised the scikit-learn \cite{pedregosa}, skorch \cite{skorch}, PyTorch \cite{pytorch}, and umap \cite{McInnes} packages. Experiments were conducted on a computer with an AMD Ryzen 3970X 32-core Processor and 32 GB RAM.

\vspace*{-2em}

\begin{table}[h!]
\centering
\caption{\label{tab:results} Comparison of reconstruction errors across DR techniques. Best and second-best models are highlighted in \textbf{\underline{bold underline}} and \textbf{bold}, respectively. Rows marked \textit{+IR} indicate results with Isotonic Regression. Values are averaged over supply and demand curves.}
\vspace{2mm} % Adds a small gap between caption and table
\footnotesize % Reduces font size for better fit
\setlength{\tabcolsep}{3.5pt} % Tightens horizontal space
\renewcommand{\arraystretch}{1.1} % Reduces row height slightly

\begin{tabular}{| l | *{2}{S[table-format=1.2, detect-weight] |} *{2}{S[table-format=1.2, detect-weight] |} *{2}{S[table-format=1.2, detect-weight] |} *{2}{S[table-format=1.2, detect-weight] |} }
\hline
& \multicolumn{2}{c|}{PCA} & \multicolumn{2}{c|}{kPCA} & \multicolumn{2}{c|}{UMAP} & \multicolumn{2}{c|}{AE} \\
\cline{2-9}
Metric & {2d} & {3d} & {2d} & {3d} & {2d} & {3d} & {2d} & {3d} \\ \hline \hline

RMSE & 4.85 & 3.48 & 5.81 & 4.53 & \textbf{\underline{2.17}} & \textbf{2.55} & 6.70 & 4.36 \\
\textit{\hspace{2mm} + IR} & \textit{4.40} & \textit{3.36} & \textit{5.27} & \textit{4.61} & \textit{2.46} & \textit{\textbf{2.55}} & \textit{6.88} & \textit{4.54} \\ \hline

MAE & 3.50 & 2.44 & 4.03 & 3.13 & \textbf{\underline{0.87}} & \textbf{1.07} & 3.31 & 2.01 \\
\textit{\hspace{2mm} + IR} & \textit{2.13} & \textit{1.66} & \textit{2.47} & \textit{2.39} & \textit{1.08} & \textit{\textbf{1.07}} & \textit{3.05} & \textit{1.88} \\ \hline

Bias & -0.08 & \textbf{\underline{-0.03}} & 0.61 & -0.12 & 0.24 & \textbf{0.04} & 0.85 & 0.45 \\
\textit{\hspace{2mm} + IR} & \textit{0.13} & \textit{-0.04} & \textit{0.64} & \textit{0.22} & \textit{0.24} & \textit{0.07} & \textit{0.62} & \textit{0.39} \\ \hline

WAPE (\%) & 23.95 & 16.73 & 27.98 & 21.90 & \textbf{\underline{6.10}} & \textbf{7.69} & 23.33 & 14.13 \\
\textit{\hspace{2mm} + IR} & \textit{15.29} & \textit{11.88} & \textit{17.50} & \textit{17.23} & \textit{7.72} & \textit{7.73} & \textit{21.57} & \textit{13.24} \\ \hline \hline

Avg. Rank & 4.50 & 2.50 & 7.00 & 5.50 & \textbf{\underline{2.00}} & \textbf{\underline{2.00}} & 8.00 & 4.50 \\ \hline
\end{tabular}
\vspace*{-3em}
\end{table}

\subsection{Preprocessing of MIBEL Data}
MIBEL operates via blind auctions where participants submit price-volume bids/asks. To handle the intractable number of pieces in these curves, we apply a preprocessing routine outlined in Section \ref{sec:Methodology}, with parameter values chosen based on preliminary experiments:
\begin{enumerate}
    \item \textbf{Price winsorization:} Prices are clipped to 99\% CI of $[0.01, 51.69]$\,EUR from train set.
    \item \textbf{Merging in prices:} We define a price grid $R_P= \{1, 2, 3,\dots, 52\}$\,EUR with $\delta_P=1$\,EUR.
    \item \textbf{Sampling in volume:} With $\delta_Q$ = 100\,MWh, we sample volumes across the 99\% CI $[20\,833, 47\,630]$\,MWh. This reduces the price dimension to $N = 268$ per curve.
\end{enumerate}

\vspace*{-1.5em}

\subsection{Dimensionality Reduction of Preprocessed MIBEL Data}
We decided to study 2d and 3d latent representations to facilitate direct visualisation. Hyperparameters were optimised via grid search on the train set:
\begin{itemize} 
    \item \textbf{kPCA:} Kernels evaluated include \{polynomial, RBF, sigmoid, cosine\}. \item \textbf{UMAP:} We tuned the number of neighbours $\{10, 11, 12, 13, 14, 15\}$, minimum distance $\{0.001, 0.01, 0.1, 0.2, 0.3, 0.4, 0.5\}$, and distance metrics \{Euclidean, Manhattan, Chebyshev\}. 
    \item \textbf{AE:} A symmetrical architecture with two hidden layers per encoder/decoder and ReLU activations, similar to \cite{sinha}. We optimised hidden unit sizes \\
    $\{[256, 128], [128, 64], [64, 32],[32, 16], [16, 8]\}$, learning rates \{0.1, 0.01, 0.001, 0.0001\}, and batch sizes $\{8, 16, 32, 64, 128\}$. 
\end{itemize}
\vspace*{-1.5em}

\subsection{Evaluation}
Model performance was assessed on the test set using RMSE, MAE, Bias, and Weighted Absolute Percentage Error (WAPE) \cite{wape}. To account for market drift, models were refit using a 7-day moving window while maintaining fixed hyperparameters from the initial grid search. Optionally, we applied IR as a post-processing step from Section \ref{sec:Methodology} to all reconstructions to enforce monotonic constraints.

\vspace*{-1em}

\begin{figure}[h!]
     \centering
     % First Subfigure: Latent Space
     \begin{subfigure}[b]{0.48\textwidth}
         \centering
         \includegraphics[width=1.03\textwidth]{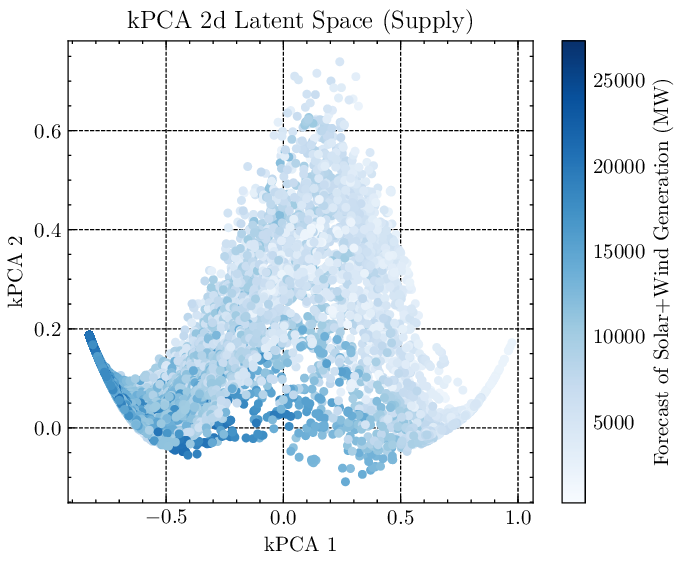}
         \caption{kPCA 2d latent space representation.}
         \label{fig:latent_space_kpca}
     \end{subfigure}
     \hfill
     % Second Subfigure: Reconstruction + Iso
     \begin{subfigure}[b]{0.48\textwidth}
         \centering
         \includegraphics[width=\textwidth]{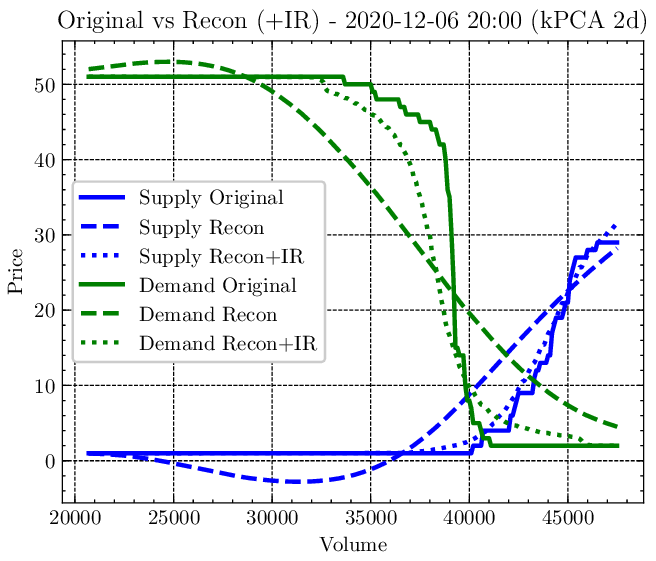}
         \caption{kPCA 2d original vs. recon. + IR.}
         \label{fig:reconstruction_kpca}
     \end{subfigure}
     
     \caption{Performance analysis of kPCA: (a) 2d latent space, (b) sample reconstruction illustrates kPCA's failure to maintain monotonicity. IR serves as a corrective layer, partially mitigating the inaccuracies that lead to results in Table~\ref{tab:results}.}
     \label{fig:kpca}
    \vspace*{-3em}
\end{figure}

\section{Results}
\vspace*{-0.5em}
The empirical outcomes of our experiments, as detailed in Table \ref{tab:results}, reveal a distinct ranking in the performance of algorithms. UMAP emerges as the best method, achieving the best average rank across error metrics in both 2d and 3d latent spaces. Expanding the latent space to 3d improved performance across most methods, with UMAP remaining the most accurate. However, increased manifold sparsity or overfitting to local noise may have decreased UMAP's gains in higher dimensions. 3d visualizations are omitted due to compactness.

\begin{figure}[h!]
     \centering
     % First Subfigure: Latent Space
     \begin{subfigure}[b]{0.48\textwidth}
         \centering
         \includegraphics[width=0.97\textwidth]{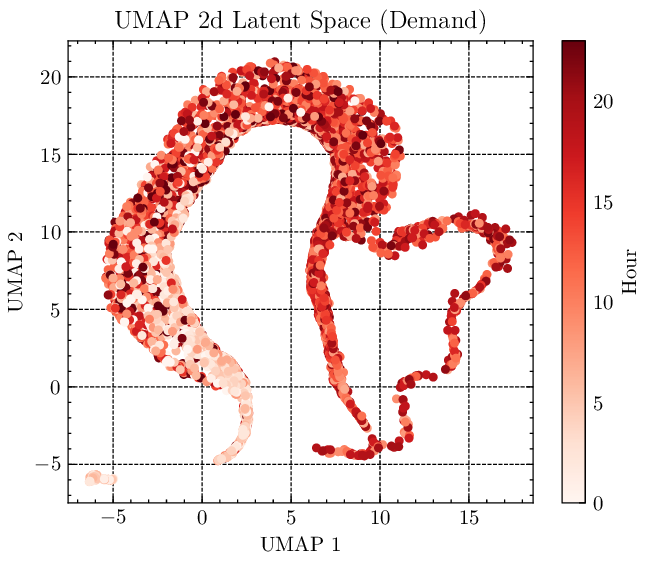}
         \caption{UMAP 2d latent space representation.}
         \label{fig:latent_space_umap}
     \end{subfigure}
     \hfill
     % Second Subfigure: Reconstruction
     \begin{subfigure}[b]{0.48\textwidth}
         \centering
         \includegraphics[width=\textwidth]{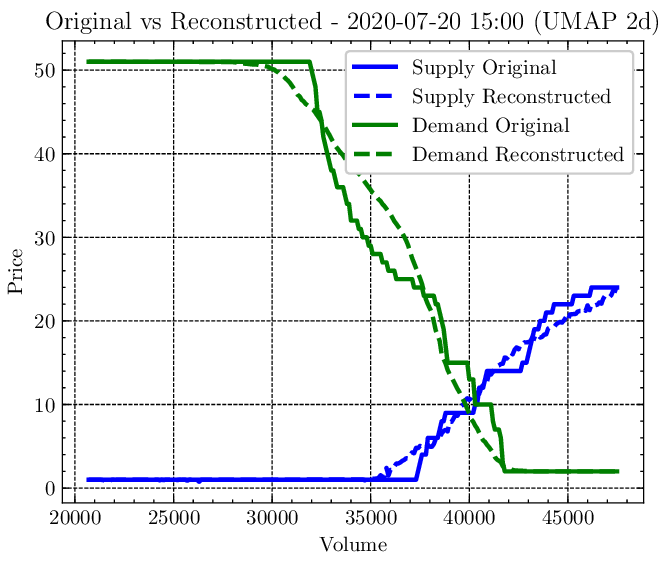}
         \caption{UMAP 2d original vs. reconstructed.}
         \label{fig:reconstruction_umap}
     \end{subfigure}
     
     \caption{Visual analysis of the top-performing model: (a) illustrates the manifold structure captured by UMAP. (b) demonstrates how UMAP preserves the monotonic nature of the task in the randomly selected summer peak.}
     \label{fig:umap}
     \vspace*{-1.5em}
\end{figure}

\begin{figure}[h!]
     \centering
     % First Subfigure: AE Latent Space
     \begin{subfigure}[b]{0.48\textwidth}
         \centering
         \includegraphics[width=0.97\textwidth]{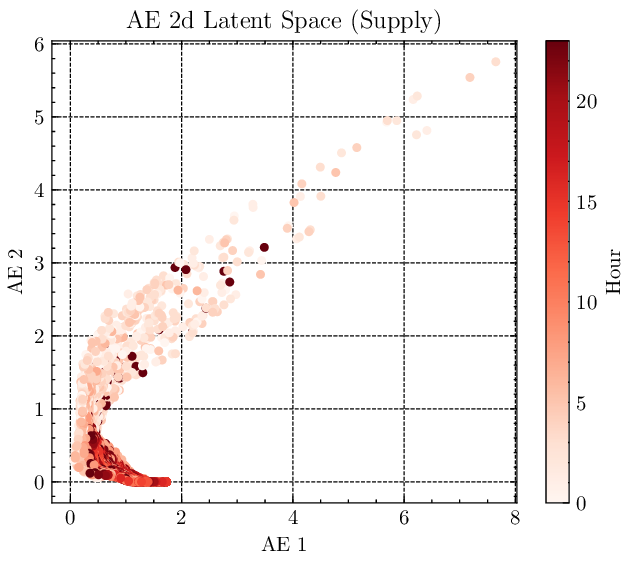}
         \caption{AE 2D latent space representation.}
         \label{fig:latent_space_ae}
     \end{subfigure}
     \hfill
     % Second Subfigure: AE Reconstruction
     \begin{subfigure}[b]{0.48\textwidth}
         \centering
         \includegraphics[width=\textwidth]{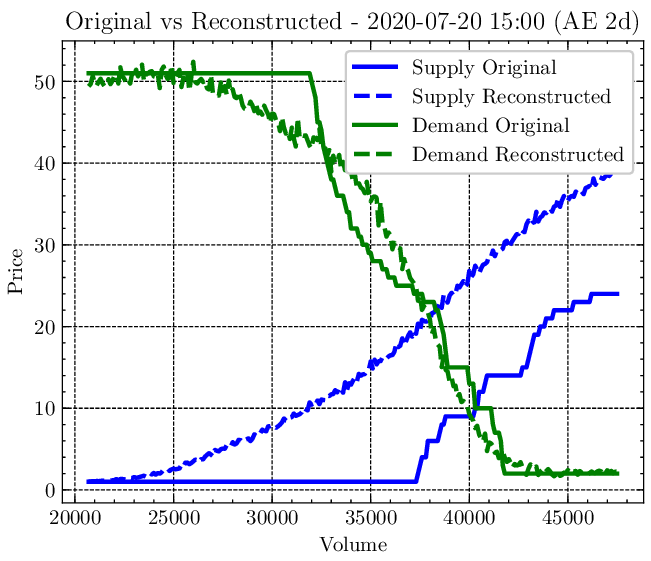}
         \caption{AE original vs. reconstructed curves.}
         \label{fig:reconstruction_ae}
     \end{subfigure}
     
     \caption{Visual analysis of the AE performance: (a) illustrates the learned latent space. (b) shows that while the AE produces visually realistic, monotonic reconstructions, it lacks the precision of other methods, resulting in higher errors.}
     \label{fig:ae_results}
     \vspace*{-2em}
\end{figure}

A visual inspection of the latent spaces (see Figures \ref{fig:latent_space_kpca},  \ref{fig:latent_space_umap}, and \ref{fig:latent_space_ae}) reveals that all models successfully capture the cyclical nature of the energy market. Beyond successfully clustering curves by the hour of the day, methods demonstrate an ability to encode broader market fundamentals, such as variations in renewable energy generation, that drive supply-side prices (see Fig. \ref{fig:latent_space_kpca}). The primary deficiency of PCA-based methods lies in the inverse transformation. As shown in Fig. \ref{fig:reconstruction_kpca}, the kPCA reconstruction often violates the fundamental structure of market curves, resulting in non-monotonic/out-of-range artefacts. Similar structural failures were observed for standard PCA; however, its visualisation is omitted for brevity. In contrast to the work by \cite{Guo}, which did not account for the monotonicity constraint, our findings suggest that PCA-based methods are insufficient for maintaining the theoretical structure of the studied market curves. This failure explains the significant performance gains observed when applying IR to impose monotonic consistency.

UMAP demonstrates an understanding of the monotonic nature of the curves. Even in the absence of post-processing, UMAP reconstructs the original curve geometry well (Fig. \ref{fig:reconstruction_umap}), preserving structures that PCA variants fail to maintain. This is reflected in the marginal improvements IR provides for UMAP, contrasting with the drastic error reductions seen in PCA and kPCA (Table \ref{tab:results}).

Regarding the AE, while the reconstructed curves appear visually realistic and generally maintain a monotonic shape (Fig. \ref{fig:reconstruction_ae}), they suffer from significantly larger error rates. This suggests that while the AE effectively learns the ``shape'' of the data, it lacks the precision in mapping achieved by UMAP, ultimately resulting in inferior accuracy.

\vspace*{-1em}
\section{Conclusions \& Future Work}
\vspace*{-0.5em}
This study demonstrates that while multiple DR methods can capture market cycles, some of them fail to preserve the functional monotonicity essential for accurate curve reconstruction. UMAP emerged as the superior architecture, securing the highest average rank across metrics by maintaining the structural integrity of the curves. We further established that IR mitigates systemic artefacts and ensures that low-dimensional representations remain economically sound.

Future research can leverage these findings for downstream predictive modelling. This includes latent-space forecasting, regime detection, to identify shifts in bidding behaviour or market anomalies. It may be valuable to explore the generalizability to other energy markets.

\vspace*{-1em}

\begin{credits}
\subsubsection{\ackname} This work was partially supported by the following projects: V4Grid: Interreg Central Europe Programme project co-funded by the European Union, project No. CE0200803; SKAI-eDIH Hopero, a project funded by European Union, GA No. 101083419. The work of Zuzana Chladn\'a was partly supported by Slovak Grant Agency APVV-20-0311. The research was conducted in consultation with ENERGY TRADING COMPANY, s.r.o. We thank them for their valuable insights and expertise.

\subsubsection{\discintname}
The authors have no competing interests to declare that are
relevant to the content of this article.
\end{credits}
%
% ---- Bibliography ----
%
% BibTeX users should specify bibliography style 'splncs04'.
% References will then be sorted and formatted in the correct style.
%
\vspace*{-1em}

\bibliographystyle{splncs04}
\bibliography{citations}

\end{document}